\documentclass[11pt]{article}

\usepackage[preprint]{acl}

\usepackage{times}
\usepackage{latexsym}

\usepackage[T1]{fontenc}
\usepackage[utf8]{inputenc}

\usepackage{microtype}

\usepackage{inconsolata}

\usepackage{graphicx}
\usepackage{amsmath}
\usepackage{cleveref}
\usepackage{amssymb}
\usepackage{algorithm,algpseudocode}
\usepackage{booktabs}
\usepackage{adjustbox}
\usepackage{xcolor}
\usepackage{tcolorbox}

\definecolor{seedblue}{RGB}{31, 108, 166}
\definecolor{refinered}{RGB}{192, 41, 43}

\newcommand{\scorecorner}[3]{%
  \makebox[3.0em][r]{#2}
  \hspace{0.15em}
  \raisebox{-0.5ex}{\scriptsize\textcolor{#1}{#3}}
}

\newcommand{\seedscore}[2]{\scorecorner{black!55}{#1}{#2}}
\newcommand{\refinescore}[2]{\scorecorner{black!55}{#1}{#2}}

\title{Each Judge Its Own Yardstick: Discovering \\ Per-VLM Taxonomies for Physical Video Evaluation}

\author{
  \textbf{Yu Cao\textsuperscript{1,\,*}},
  \textbf{Ziquan Liu\textsuperscript{1}},
  \textbf{Zhensong Zhang\textsuperscript{2}},
  \textbf{Jiankang Deng\textsuperscript{3}},
  \textbf{Shaogang Gong\textsuperscript{1}},
  \textbf{Jifei Song\textsuperscript{2}}
\\
  \textsuperscript{1}Queen Mary University of London \quad
  \textsuperscript{2}Huawei Darwin Research Center \quad
  \textsuperscript{3}Imperial College London
\\
  \textsuperscript{*}Corresponding author
\\
  \texttt{\{yu.cao, ziquan.liu, s.gong\}@qmul.ac.uk}
\\
  \texttt{\{zhangzhensong, jifeisong\}@huawei.com, j.deng@imperial.ac.uk}
}

\begin{document}
\maketitle

\begin{abstract}
Maintaining physical consistency in video generators and world models increasingly relies on vision-language models (VLMs) as automated judges that provide reward signals, ranking decisions, and data-filtering criteria. Yet VLMs differ substantially in training data and architecture, encoding physical phenomena through distinct internal representations. A single global evaluation schema therefore gives every VLM the same axes of competence, regardless of what each can actually perceive. We propose \textsc{JudgeFit}, an iterative refinement procedure that discovers a per-VLM evaluation taxonomy. An initial taxonomy is constructed by prompting the target VLM to enumerate physics errors on a small set of videos and clustering the resulting descriptions. The taxonomy is then refined through a diagnostic step: we calibrate the VLM's per-dimension scores to human physical-commonsense ratings, diagnose which dimensions it scores unreliably or redundantly, and prompt an LLM to repair them, iterating until convergence. We further instantiate this procedure as a benchmark and apply it to 16 VLMs spanning eight model families. The refined taxonomy outperforms the global-schema baseline on held-out videos for every VLM tested, with a mean relative improvement of approximately 32\%. Beyond aggregate accuracy, the per-VLM profiles expose model-specific blind spots that overall rankings cannot anticipate, with reliability patterns differing markedly across model families.
\end{abstract}

\section{Introduction}

Recent video and world models~\citep{brooks2024sora,yang2025cogvideox,wan2025wan,agarwal2025cosmos} increasingly rely on vision-language models (VLMs) as automated judges, extending the LLM-as-judge paradigm \citep{zheng2023judging,gu2024survey} to the multimodal setting. These judges supply reward signals for reinforcement learning \citep{christiano2017deep, ouyang2022training, rocamonde2024vlm}, train large-scale visual reward models for text-to-image and text-to-video generation \citep{xu2023imagereward, wu2025rewarddance, liu2025improving}, and grade outputs on video-quality and physical-consistency benchmarks \citep{he2024videoscore, bansal2025videophy, meng2024towards}. The reliability of these judges directly bounds the quality of any downstream generative system trained or filtered against them \citep{gao2023scaling}. 

\begin{figure}[t]
  \includegraphics[width=\columnwidth]{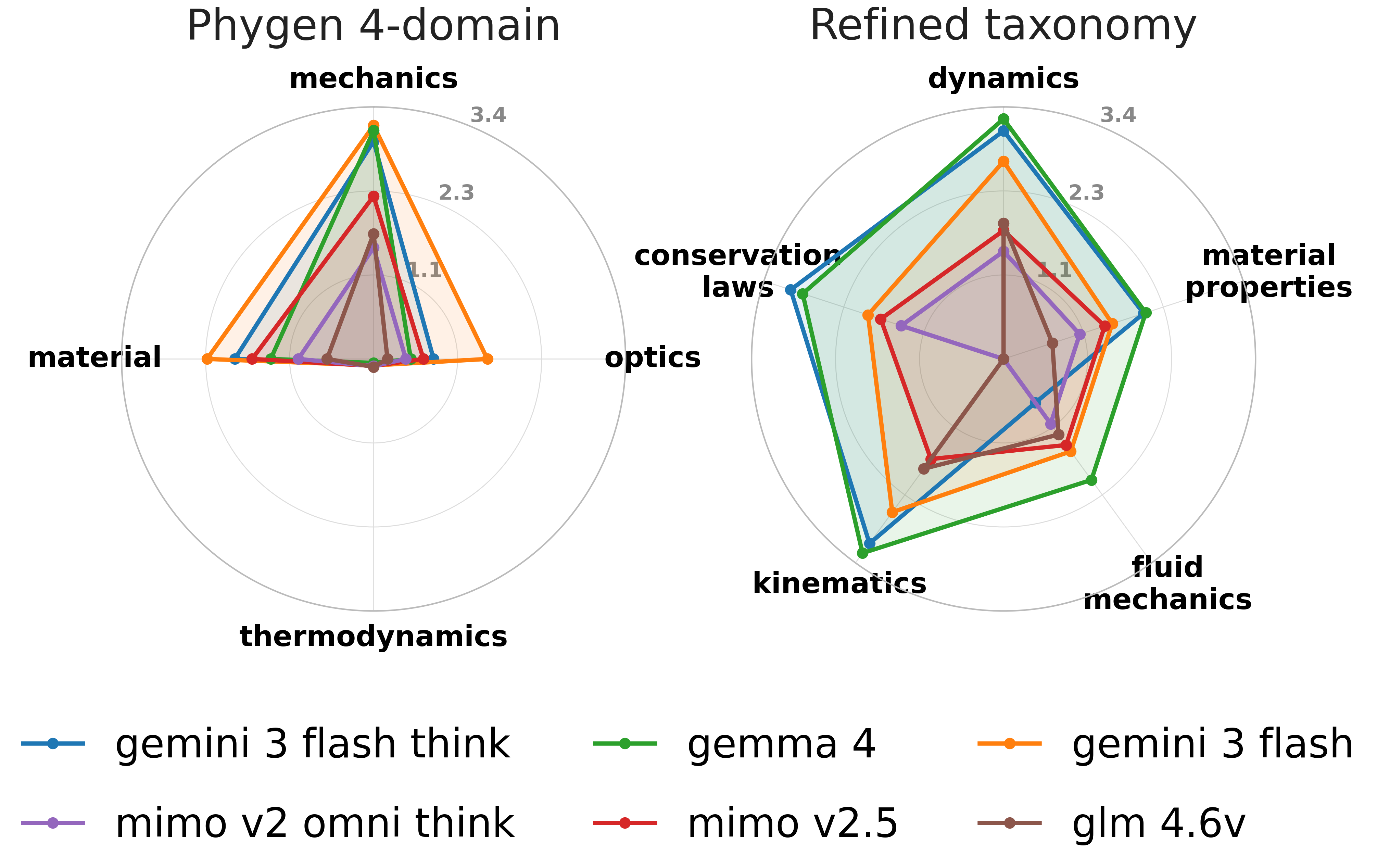}
  \vspace{-2 em}
  \caption{\textbf{Evaluation under two taxonomies.} The fixed four-domain schema (left) concentrates VLM scores on the mechanics axis with the other three axes nearly inactive, offering little resolution on cross-model differences. Our per-VLM refined taxonomy (right) renders every dimension discriminative and reveals distinct strengths across models.}
  \vspace{-1.5 em}
  \label{fig:motivation}
\end{figure}

On physical commonsense in particular, multimodal judges have been shown to disagree substantially with human raters and to exhibit position bias and scoring instability \citep{chen2024mllm,wang2024fair}. Existing physical-consistency benchmarks~\citep[e.g.,][]{meng2024towards,bansal2025videophy2} respond by prescribing a fixed evaluation schema and applying it uniformly across all VLMs. 
This design implicitly assumes that VLMs share the same axes of physical-evaluation competence and differ only in magnitude along each \citep{yin2024survey}. As shown in Figure~\ref{fig:motivation}, under PhyGenBench's four-domain schema~\citep{meng2024towards}, scores from six VLMs concentrate on the mechanics axis, leaving the remaining three nearly inactive, whereas our refined per-VLM taxonomy yields a distinct profile for each model in which overall standing does not predict competence on any particular rule.

We argue that the evaluation schema should be treated as a property of the evaluator rather than of the benchmark, and that each VLM should be assessed along dimensions it can reliably score. Pursuing this view requires a procedure that discovers a taxonomy per evaluator, and yields, as a direct by-product, a per-model map of which physical rules each VLM can and cannot ground in human judgment.

We introduce \textsc{JudgeFit}, a two-stage seed-and-refine pipeline that discovers a per-VLM evaluation taxonomy.
The seed stage has the target VLM enumerate, in free-form natural language, the physics errors it sees in a small set of videos stratified across the human quality spectrum. An LLM then clusters these descriptions into an initial taxonomy. The resulting taxonomy is grounded in the VLM's own perceptual vocabulary: concepts the VLM never raises are concepts it cannot reliably detect, and are correspondingly absent from its starting taxonomy.
In the refine stage, we calibrate the VLM's per-dimension scores against human physical-commonsense ratings and use this to diagnose the taxonomy's weaknesses: unreliable dimensions, redundant ones, and error classes no dimension captures. An LLM then proposes local edits, keeping only those whose dimensions improve agreement with human judgment beyond an overall sense of how flawed a video looks, iterating until convergence.
Our refinement loop stabilizes within roughly two rounds, whereas scalar-reward search procedures \citep{yang2024opro,yuksekgonul2025optimizing} require hundreds of iterations to recover per-dimension information from aggregate scores and become impractical when each call to a VLM is far more expensive than a text-only LLM query.

We apply this procedure to 16 VLMs spanning eight model families, ranging from 7B open-weight models to frontier closed-source systems, using human ratings from VideoPhy-2 \citep{bansal2025videophy2} as the alignment target. The refined per-VLM taxonomy outperforms the baseline on held-out videos for every VLM tested, with a mean relative improvement of approximately 32\%. The gain holds across the full capability spectrum, and the resulting per-VLM profiles reveal that strong-on-average VLMs are not uniformly strong across physical rules, with the rules a given model judges reliably varying substantially across model families.
Our contributions are as follows:
\begin{itemize}
    \item \textbf{\textsc{JudgeFit}}, a seed-and-refine pipeline that treats the judge's evaluation schema itself as an optimizable object, using a calibrated diagnosis of which dimensions a VLM can reliably and distinctly score to drive sample-efficient edits. The procedure converges within roughly two refinement rounds in our experiments.
    \item \textbf{A per-VLM benchmark study} on VideoPhy-2 covering 16 VLMs across eight model families, releasing the refined per-VLM taxonomies and held-out evaluations.
    \item \textbf{A cross-VLM analysis} of concept convergence and a bias-by-discrimination typology over physical rules, identifying which physics concepts are universally legible across VLMs and which remain model-specific blind spots.
\end{itemize}

\begin{figure*}[t]
  \includegraphics[width=\textwidth]{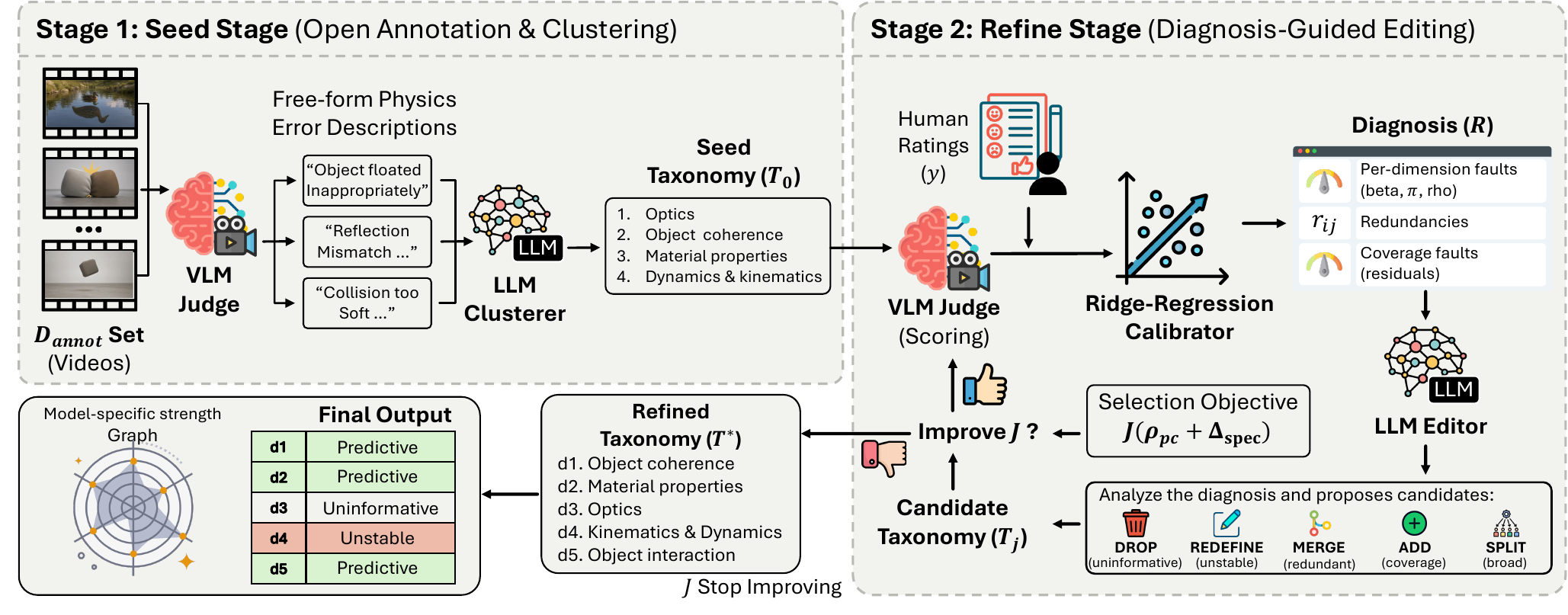}
  \vspace{-2 em}
  \caption{\textbf{Overview of \textsc{JudgeFit}.} \emph{Stage 1 (Seed)}: the VLM judge describes physics errors in free-form text over the annotation set $\mathcal{D}_{\text{annot}}$, and an LLM clusters these descriptions into the initial seed taxonomy $\mathcal{T}_0$. \emph{Stage 2 (Refine)}: the VLM scores each video along the current dimensions, a ridge-regression calibrator maps these scores to human ratings $y$, and the resulting diagnosis $R$ drives an LLM editor that proposes local edits (drop, redefine, merge, add, split). A candidate is kept only if it improves the selection objective $J$, and the loop repeats until no edit helps, yielding the refined taxonomy $\mathcal{T}^\star$ in which each dimension tend to be predictive, stable, and non-redundant.}
  \vspace{-1 em}
  \label{fig:pipeline}
\end{figure*}

\section{Related Work}
\noindent\textbf{VLMs as automated judges.}
The LLM-as-judge paradigm \citep{zheng2023judging} has been extended to multimodal settings, where vision-language models score generated images and videos against human preferences \citep{chen2024mllm,he2024videoscore}. VLM-based judges supply reward signals for reinforcement learning \citep{rocamonde2024vlm}, train large-scale visual reward models for text-to-image and text-to-video generation \citep{wu2025rewarddance, liu2025improving}, and serve as automated graders in video-quality benchmarks \citep{he2024videoscore}. However, multimodal judges show substantial disagreement with human raters and exhibit position bias, scoring instability, and hallucination \citep{chen2024mllm, wang2024fair, shi2025judging}. Judges also favor outputs from their own model family \citep{panickssery2024llm, wataoka2024selfpreference}. The dominant response has been to calibrate or debias judges toward a shared rubric \citep{wang2024fair, liu2024calibrating, zheng2023judging}, treating evaluator disagreement as noise to be removed rather than signal about what each model can reliably score.

\noindent\textbf{Physical-consistency benchmarks for video generation.}
VideoPhy \citep{bansal2025videophy} evaluates semantic adherence and physical commonsense across 688 prompts, and VideoPhy-2 \citep{bansal2025videophy2} extends this to 3{,}940 action-centric prompts with per-video physical-rule annotations. PhyGenBench \citep{meng2024towards} organizes 160 prompts under 27 physical laws spanning four fixed domains and applies a shared evaluation pipeline (PhyGenEval) across all judges. Adjacent benchmarks probe physical reasoning of VLMs themselves \citep{chow2025physbench,xiang2025seephys} or measure pixel-level physics fidelity in generated videos \citep{motamed2026generative}. Across these benchmarks, evaluation criteria are fixed in advance and applied uniformly across all judges.

\noindent\textbf{LLM-driven optimization and rubric induction.}
Two adjacent lines of work inform our refinement procedure. The first uses LLMs as iterative optimizers over textual artifacts, including prompts \citep{yang2024opro,pryzant2023automatic} and compound-system components \citep{yuksekgonul2025optimizing,khattab2023dspy}, with OPRO \citep{yang2024opro} notably finding that optimal prompts diverge across model families, an early signal that per-evaluator adaptation can outperform shared artifacts. The second induces or refines evaluation criteria with LLM assistance: EvalGen \citep{shankar2024who} and AutoCalibrate \citep{liu2024calibrating} construct rubrics from feedback, LLM-Rubric \citep{hashemi2024llmrubric} fits a calibrator over a human-defined multidimensional rubric for personalized score prediction, and GER-Eval \citep{siro2026learning} reports that LLM-generated rubrics fragment across model families. These works target LLM text evaluation with human-defined or one-shot induced rubrics. Our setting differs along three axes: the evaluator is a VLM judging video physics rather than an LLM judging text \citep{meng2024towards,bansal2025videophy2}; the taxonomy is iteratively discovered per evaluator rather than fixed or generated in one shot; and a linear calibrator over human ratings is used as a diagnostic signal that steers refinement rather than as a final score predictor.
These works also operate in regimes where evaluator calls are cheap text-only LLM queries; transferring iterative refinement to VLM video evaluation, where each call ingests sampled frames, places a tighter budget on the search procedure and motivates a sample-efficient design.

\section{The JudgeFit Pipeline}
\label{sec:pipeline}

\subsection{Overview and Notation}
\label{sec:overview}
\textsc{JudgeFit} discovers, for a given VLM judge $v$, an evaluation taxonomy tailored to what $v$ can reliably perceive. A taxonomy $\mathcal{T} = \{(d_1, \delta_1), \dots, (d_k, \delta_k)\}$ is a set of $k$ named physical-error dimensions, each $d_i$ paired with a natural-language definition $\delta_i$. Given a video $x$, the judge scores every dimension for error severity, producing a vector $\mathbf{s}^{v}_{\mathcal{T}}(x) \in [0,5]^{k}$. We assess a taxonomy by how well these per-dimension scores explain human physical-commonsense ratings $y(x) \in \{1,\dots,5\}$.

The procedure has two stages (Figure~\ref{fig:pipeline}). A \emph{seed} stage (\S\ref{sec:seed}) elicits an initial taxonomy from the judge's own error descriptions on an annotation set $\mathcal{D}_{\text{annot}}$, stratified across the full quality spectrum from physically faithful to severely implausible videos. A \emph{refine} stage (\S\ref{sec:refine}) then tunes the taxonomy's per-dimension scores toward the human ratings on $\mathcal{D}_{\text{annot}}$. A disjoint test set $\mathcal{D}_{\text{test}}$, never seen during refinement, is held out for evaluation.

\subsection{Seed: Open Annotation and Clustering}
\label{sec:seed}
For each video in $\mathcal{D}_{\text{annot}}$, we prompt $v$ to list, in free-form text, the physics errors it observes without predefined dimensions. An LLM then clusters the pooled descriptions into named, defined categories, which form the seed taxonomy $\mathcal{T}_0$. Because $\mathcal{T}_0$ is built only from errors the judge reports on its own, its dimensions are exactly those the judge already attends to, and an error class it never mentions yields no dimension. The seed is thus specific to $v$ before any refinement begins.

\subsection{Refine: Diagnosis-Guided Editing}
\label{sec:refine}
The seed captures the errors a VLM notices, but downstream uses such as supplying reward signals or ranking generations demand more than a list of noticed errors: they require per-dimension scores that track human judgment, which the seed does not yet provide. Refinement repairs the taxonomy through a loop in which an LLM editor revises the dimensions while the VLM re-scores the videos along them.

The two models play deliberately asymmetric roles. The LLM editor can see how humans rated each video and which physical rules each one violated, but can influence the VLM only by revising the taxonomy. The VLM, in turn, scores each video using the current taxonomy, with no access to the human ratings. Because the editor reaches the VLM only through the taxonomy, every score remains the VLM's own, and agreement with human ratings can improve only when a revision gives the VLM dimensions it is actually able to grade.

Each round, the editor receives a \textsc{Diagnose} (\S\ref{sec:diagnosis}) of where the current taxonomy fails and proposes local, evidence-backed edits in response. 
The VLM re-scores the videos under each candidate, and a candidate is accepted only if it improves a selection objective $J$ (Eq.~\ref{eq:objective}) that rewards a taxonomy not just for ranking videos by human physical commonsense but for doing so through genuinely distinct dimensions. 
When no edit improves $J$, refinement stops. Empirically, most of the gain arrives within roughly two rounds (Figure~\ref{fig:ranking}). This efficiency matters because each candidate sends the VLM back to re-score every video, far costlier than the text-only queries that scalar-reward optimizers iterate over hundreds of times \citep{yang2024opro,yuksekgonul2025optimizing}.

\begin{algorithm}[t]
\caption{Taxonomy refinement procedure}
\label{alg:refine}
\begin{algorithmic}[1]
\Require Seed taxonomy $\mathcal{T}_0$, annotation set $\mathcal{D}_{\text{annot}}$ with ratings $y$ and free-form annotations $\mathcal{A}$, VLM $v$, LLM editor $\mathcal{E}$, budget $T_{\max}$
\Ensure Refined taxonomy $\mathcal{T}^\star$, calibrator $f^\star$
\State $\mathbf{S}_0 \gets \textsc{Score}(v, \mathcal{T}_0, \mathcal{D}_{\text{annot}})$
\State $(J^\star, f^\star) \gets \textsc{Select}(\mathbf{S}_0, y)$
\State $\mathcal{T}^\star, \mathbf{S}^\star \gets \mathcal{T}_0, \mathbf{S}_0$;\quad $t \gets 0$
\State $\textit{improved} \gets \textbf{true}$
\While{$\textit{improved}$ \textbf{and} $t < T_{\max}$}
    \State $\textit{improved} \gets \textbf{false}$;\quad $t \gets t+1$
    \State $R \gets \textsc{Diagnose}(\mathbf{S}^\star, y, \mathcal{A})$
    \State $\{\mathcal{T}_j\}_{j=1}^{m} \gets \mathcal{E}(\mathcal{T}^\star, R)$ \Comment{local edits}
    \State \textbf{for each} $j$:\ $(\mathbf{S}_j, J_j, f_j) \gets \textsc{Eval}(\mathcal{T}_j)$
    \State $j^\star \gets \arg\max_j J_j$
    \If{$J_{j^\star} > J^\star$} \Comment{accept the edit}
        \State $\mathcal{T}^\star, \mathbf{S}^\star, J^\star, f^\star \gets \mathcal{T}_{j^\star}, \mathbf{S}_{j^\star}, J_{j^\star}, f_{j^\star}$
        \State $\textit{improved} \gets \textbf{true}$
    \EndIf
\EndWhile
\State \Return $\mathcal{T}^\star, f^\star$
\end{algorithmic}
\end{algorithm}

\section{Diagnostic Signals}
\label{sec:diagnosis}
To edit the taxonomy without leaking answers, the editor needs a precise account of how the VLM's dimension scores relate to human ratings. We fit a cross-validated, ridge-regularized linear map $f_{\mathcal{T}}$ from a video's per-dimension scores $\mathbf{s}(x)$ to its human rating $y(x)$, giving a calibrated prediction $\hat{y}_{\text{full}} = f_{\mathcal{T}}(\mathbf{s})$, and read the taxonomy's faults off this map. Calibrating, rather than comparing scores directly, lets us judge a dimension by whether it carries information about human ratings, independent of how each VLM uses the scale.

\subsection{What the diagnosis reveals}

\noindent\textbf{Per-dimension faults.}
For each dimension $i$ we read three signals from the calibrator: its standardized coefficient $\beta_i$, the consistency $\pi_i \in [0,1]$ of that coefficient's sign across folds, and its agreement with human ratings,
\begin{equation}
\rho_i = \frac{\mathrm{cov}\big(\mathrm{rk}(s_i),\, \mathrm{rk}(y)\big)}{\sigma_{\mathrm{rk}(s_i)}\,\sigma_{\mathrm{rk}(y)}},
\label{eq:spearman}
\end{equation}
the Spearman rank correlation between dimension $i$'s scores and the human rating, where $\mathrm{rk}(\cdot)$ is the rank transform over videos.
Applying the same correlation between two dimensions gives $r_{ij}$, which is large when the VLM scores them interchangeably.
The diagnosis flags faults of two kinds, each implying a repair drawn from five operations. 

\noindent The first kind concerns a single dimension:
\vspace{-0.5 em}
\begin{equation}
\begin{cases}
\;|\rho_i| < \tau & \Rightarrow\;\textsc{drop} \quad (\text{uninformative}) \\[4pt]
\;\pi_i < \kappa & \Rightarrow\;\textsc{redefine} \quad (\text{unstable}) \\[4pt]
\;\displaystyle\max_{j\neq i} r_{ij} > \gamma & \Rightarrow\;\textsc{merge} \quad (\text{redundant})
\end{cases}
\label{eq:faults}
\end{equation}
An \emph{uninformative} dimension's scores rank videos no better than chance against human judgment, marking a distinction the VLM cannot perceive. An \emph{unstable} one contributes to the prediction with a sign that flips across folds, marking a distinction the VLM applies inconsistently. A \emph{redundant} pair ranks videos almost identically, so the VLM does not truly separate them and the calibrator cannot credit either.

\noindent\textbf{Coverage faults.}
A complementary fault hides in the videos $f_{\mathcal{T}}$ predicts worst. When $\hat{y}_{\text{full}}(x)$ exceeds $y(x)$, the VLM missed errors humans penalized; when it falls short, the VLM over-flagged errors humans tolerated. Aligning these residuals with the VLM's own free-form annotations of the same videos reveals error classes that recur yet that no dimension captures, prompting the editor to \textsc{add} a dimension or \textsc{split} an overly broad one.

\begin{table*}[t]
\centering
\small
\setlength{\tabcolsep}{8pt} 
\renewcommand{\arraystretch}{1.15}
% \caption{Per-VLM agreement with human ratings (test-set Spearman $\rho$) under the fixed \textsc{PhyGen} baseline, \textsc{Seed}, and \textsc{Refine}. Differences from \textsc{PhyGen} shown at the lower-right of each \textcolor{seedblue}{Seed} and \textcolor{refinered}{Refine} score. Thinking variants marked with $\dagger$.}
\caption{Per-VLM agreement with human ratings (test-set Spearman $\rho$) under the fixed \textsc{PhyGen} baseline, \textsc{Seed}, and \textsc{Refine}. Differences from \textsc{PhyGen} shown at the lower-right of each \textsc{Seed} and \textsc{Refine} score. Thinking variants are marked with $\dagger$.}
\vspace{-1 em}
\label{tab:vlm_results}
\begin{adjustbox}{max width=\textwidth}
\begin{tabular}{lccc}
\toprule
\textbf{Model} & \textbf{PhyGen} & \textbf{Seed} & \textbf{Refine} \\
\midrule
\multicolumn{4}{l}{\textit{Gemini}} \\
Gemini-3-Flash-Thinking$^\dagger$~\citep{google2025gemini3flash} & +0.346 & \seedscore{+0.372}{+0.026} & \refinescore{+0.484}{+0.138} \\
Gemini-3-Flash~\citep{google2025gemini3flash}                    & +0.293 & \seedscore{+0.400}{+0.107} & \refinescore{+0.429}{+0.136} \\
\midrule
\multicolumn{4}{l}{\textit{Seed}} \\
Seed-2.0-Lite-Thinking$^\dagger$~\citep{seed2026seed2} & +0.310 & \seedscore{+0.400}{+0.091} & \refinescore{+0.422}{+0.112} \\
Seed-2.0-Lite~\citep{seed2026seed2}                    & +0.261 & \seedscore{+0.286}{+0.025} & \refinescore{+0.308}{+0.047} \\
\midrule
\multicolumn{4}{l}{\textit{MiMo}} \\
MiMo-v2.5-Thinking$^\dagger$~\citep{xiao2026mimov2flash}    & +0.247 & \seedscore{+0.356}{\textbf{+0.109}} & \refinescore{+0.406}{\textbf{+0.159}} \\
MiMo-v2.5~\citep{xiao2026mimov2flash}                       & +0.317 & \seedscore{+0.305}{-0.012} & \refinescore{+0.330}{+0.012} \\
MiMo-v2-Omni~\citep{xiao2026mimov2flash}                    & +0.294 & \seedscore{+0.341}{+0.046} & \refinescore{+0.319}{+0.025} \\
MiMo-v2-Omni-Thinking$^\dagger$~\citep{xiao2026mimov2flash} & +0.234 & \seedscore{+0.285}{+0.051} & \refinescore{+0.291}{+0.057} \\
\midrule
\multicolumn{4}{l}{\textit{Qwen}} \\
Qwen2.5-VL-7B~\citep{bai2025qwen25vl}        & +0.317 & \seedscore{+0.276}{-0.041} & \refinescore{+0.383}{+0.066} \\
Qwen3-VL-8B-Instruct~\citep{qwen2025qwen3vl} & +0.307 & \seedscore{+0.328}{+0.021} & \refinescore{+0.375}{+0.068} \\
\midrule
\multicolumn{4}{l}{\textit{Gemma}} \\
Gemma-4-31B~\citep{gemma4_2026} & +0.285 & \seedscore{+0.355}{+0.070} & \refinescore{+0.371}{+0.086} \\
\midrule
\multicolumn{4}{l}{\textit{Nova}} \\
Nova-2-Lite~\citep{amazon2025nova2} & +0.215 & \seedscore{+0.174}{-0.041} & \refinescore{+0.262}{+0.047} \\
\midrule
\multicolumn{4}{l}{\textit{GLM}} \\
GLM-5V-Turbo~\citep{zai2026glm5v}             & +0.181 & \seedscore{+0.194}{+0.013} & \refinescore{+0.207}{+0.026} \\
GLM-4.6V-Thinking$^\dagger$~\citep{glmv2025} & +0.037 & \seedscore{+0.136}{+0.100} & \refinescore{+0.163}{+0.126} \\
GLM-4.6V~\citep{glmv2025}                     & +0.116 & \seedscore{+0.132}{+0.016} & \refinescore{+0.138}{+0.022} \\
\midrule
\multicolumn{4}{l}{\textit{Reka}} \\
Reka-Edge~\citep{ormazabal2024reka} & +0.070 & \seedscore{+0.043}{-0.028} & \refinescore{+0.151}{+0.081} \\
\bottomrule
\end{tabular}
\end{adjustbox}
\vspace{-0.5 em}
\end{table*}

\subsection{The Selection Objective}
\label{sec:selection}
The two fault types tell the editor what to change, but not whether a change is worth keeping. That decision must guard against a taxonomy that improves agreement with humans by collapsing into a single impression of overall badness, ranking videos well while revealing nothing about \emph{which} physics failed. We therefore compare the full calibrator against a one-feature \emph{halo} baseline~\citep{thorndike1920constant} $\hat{y}_{\text{halo}}$ fit on the mean dimension score $\bar{s} = \tfrac{1}{k}\sum_i s_i$ alone, and credit a taxonomy only for the ranking its dimensions buy beyond this baseline. Reusing the rank correlation $\rho(\cdot)$ of Eq.~\eqref{eq:spearman} between a prediction and the human ratings, the refinement objective is:
\vspace{-1 em}
\begin{equation}
J(\mathcal{T}) \;=\; \underbrace{\rho(\hat{y}_{\text{full}})}_{\text{ranking quality}} \;+\; \underbrace{\big[\,\rho(\hat{y}_{\text{full}}) - \rho(\hat{y}_{\text{halo}})\,\big]}_{\Delta_{\text{spec}}:\ \text{specificity gain over halo}} .
\label{eq:objective}
\end{equation}
An edit that improves $\rho(\hat{y}_{\text{full}})$ only by collapsing the taxonomy toward a single overall-quality score gains no specificity, and is rejected by $J$.
% A persistently small $\Delta_{\text{spec}}$ is then not a failed search but a measurement: this VLM grades physics on one axis of overall plausibility, and no taxonomy can make it resolve more.

\section{Experiments}

\noindent\textbf{Data and Baseline.}
We align against VideoPhy-2 \citep{bansal2025videophy2}, using its per-video human ratings (1--5) and physical-rule annotations. We sample 300 videos balanced across the five rating levels, split into a 200-video annotation set $\mathcal{D}_{\text{annot}}$ for seeding and refinement and a disjoint 100-video test set $\mathcal{D}_{\text{test}}$ for evaluation. Our fixed-schema baseline (\textsc{PhyGen}) scores every judge along the four physical domains of PhyGenBench \citep{meng2024towards} under the same calibration protocol, so it differs from \textsc{Refine} only in its dimensions.\\

\noindent\textbf{Models.}
We evaluate VLMs that natively accept video input, so each judge sees the full clip rather than a sampled subset of frames. This yields 16 VLMs across eight model families, covering both open-weight (Qwen, Gemma, GLM, Reka) and closed-source (Gemini, Seed, Nova, MiMo) systems, with thinking and non-thinking variants included where available. The full list of VLMs with citations and the implementation details is given in \Cref{tab:vlm_results} and \Cref{app:impl}. The LLM editor and clustering model are fixed to Gemini-2.5-Pro \citep{geminiteam2025gemini25} across all judges, chosen for its long context window, which the clustering step requires to pool free-form error descriptions over the full annotation set.

\subsection{Refinement Improves Every VLM}
\label{sec:main_results}

Table~\ref{tab:vlm_results} reports test-set agreement under three variants. \textsc{Refine} outperforms the \textsc{PhyGen} baseline for all 16 VLMs, raising mean correlation from 0.239 to 0.315, a relative gain of 32\%.
The improvement holds across the full capability spectrum: the strongest judge (Gemini-3-Flash-Thinking, 0.346) and the weakest (GLM-4.6V, 0.116) both benefit, with the largest gains from MiMo-v2.5-Thinking ($+0.159$) and Gemini-3-Flash-Thinking ($+0.138$).

\begin{table}[th]
\centering
\small
\renewcommand{\arraystretch}{1.2}
\caption{Effect of a thinking mode under \textsc{Seed} and \textsc{Refine}. Green and red arrows indicate performance improvement and degradation respectively.}
\vspace{-1 em}
\label{tab:thinking}
\begin{adjustbox}{max width=\columnwidth}
\begin{tabular}{lcc@{\hspace{2em}}cc}
\toprule
\noalign{\vskip -0.4em}
& \multicolumn{2}{c}{\textbf{Seed}} & \multicolumn{2}{c}{\textbf{Refine}} \\[-0.35em]
\cmidrule(lr){2-3}\cmidrule(lr){4-5}
\noalign{\vskip -0.35em}
\textbf{Model} & \textbf{Base} & \textbf{+Think} & \textbf{Base} & \textbf{+Think} \\[-0.35em]
\midrule
\noalign{\vskip 0.1em}
Gemini-3-Flash & 0.400 & 0.372 \textcolor{red}{$\downarrow$} & 0.429 & \textbf{0.484} \textcolor{green}{$\uparrow$} \\
Seed-2.0-Lite  & 0.286 & 0.400 \textcolor{green}{$\uparrow$} & 0.308 & \textbf{0.422} \textcolor{green}{$\uparrow$} \\
MiMo-v2.5      & 0.305 & 0.356 \textcolor{green}{$\uparrow$} & 0.330 & \textbf{0.406} \textcolor{green}{$\uparrow$} \\
MiMo-v2-Omni   & \textbf{0.341} & 0.285 \textcolor{red}{$\downarrow$} & 0.319 & 0.291 \textcolor{red}{$\downarrow$} \\
GLM-4.6V       & 0.132 & 0.136 \textcolor{green}{$\uparrow$} & 0.138 & \textbf{0.163} \textcolor{green}{$\uparrow$} \\
\bottomrule
\end{tabular}
\end{adjustbox}
\vspace{-3 em}
\end{table}

The size of the gain varies by family rather than tracking baseline strength. The Gemini and GLM families improve consistently under refinement, with GLM-4.6V-Thinking and Gemini-3-Flash-Thinking gaining $+0.126$ and $+0.138$ over baseline, and Reka-Edge, the weakest seed, recovered to a competitive level once refined ($+0.081$). The MiMo family gains the least overall, with three of its four variants improving by under $+0.06$, indicating that some judges expose little headroom for a refined taxonomy to exploit.\\

\noindent\textbf{Thinking variants.} We also compare thinking and non-thinking variants under \textsc{Refine} (Table~\ref{tab:thinking}). Four of the five families improve with a thinking mode and one (MiMo-v2-Omni) declines, and the effect does not track base capability: three families with near-identical base scores (Seed-2.0-Lite $0.308$, MiMo-v2-Omni $0.319$, MiMo-v2.5 $0.330$) respond in different directions and magnitudes ($+0.114$, $-0.028$, $+0.076$). Unlike the broadly consistent improvements a thinking mode tends to bring in text-only LLM reasoning, its benefit for physical video judging is model-specific rather than systematic.

\begin{figure}[th]
  \includegraphics[width=\columnwidth]{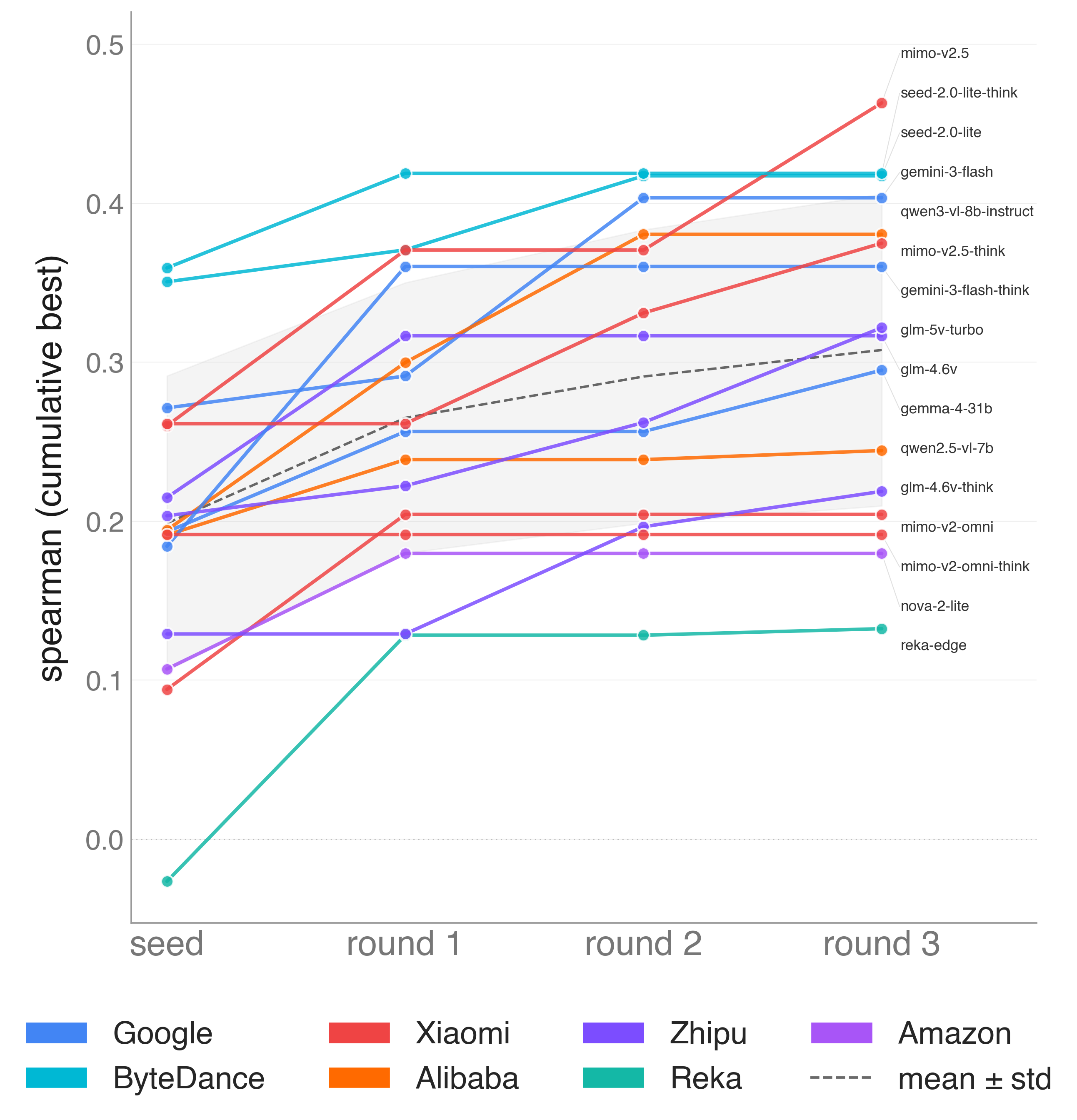}
  \vspace{-2 em}
  \caption{\textbf{Refinement converges quickly and lifts weak seeds.} Selection objective on $\mathcal{D}_{\text{annot}}$ across refinement rounds, colored by developer (cumulative best per model). The mean trajectory (dashed, $\pm$ std) rises most steeply from seed to round~1 and flattens thereafter, with most models plateauing by round~2. Reka-Edge starts below zero at the seed and gains the most from a single round, while strong seeds (ByteDance) move little.}
  \vspace{-1.5 em}
  \label{fig:ranking}
\end{figure}

\begin{figure*}[t]
  \includegraphics[width=\textwidth]{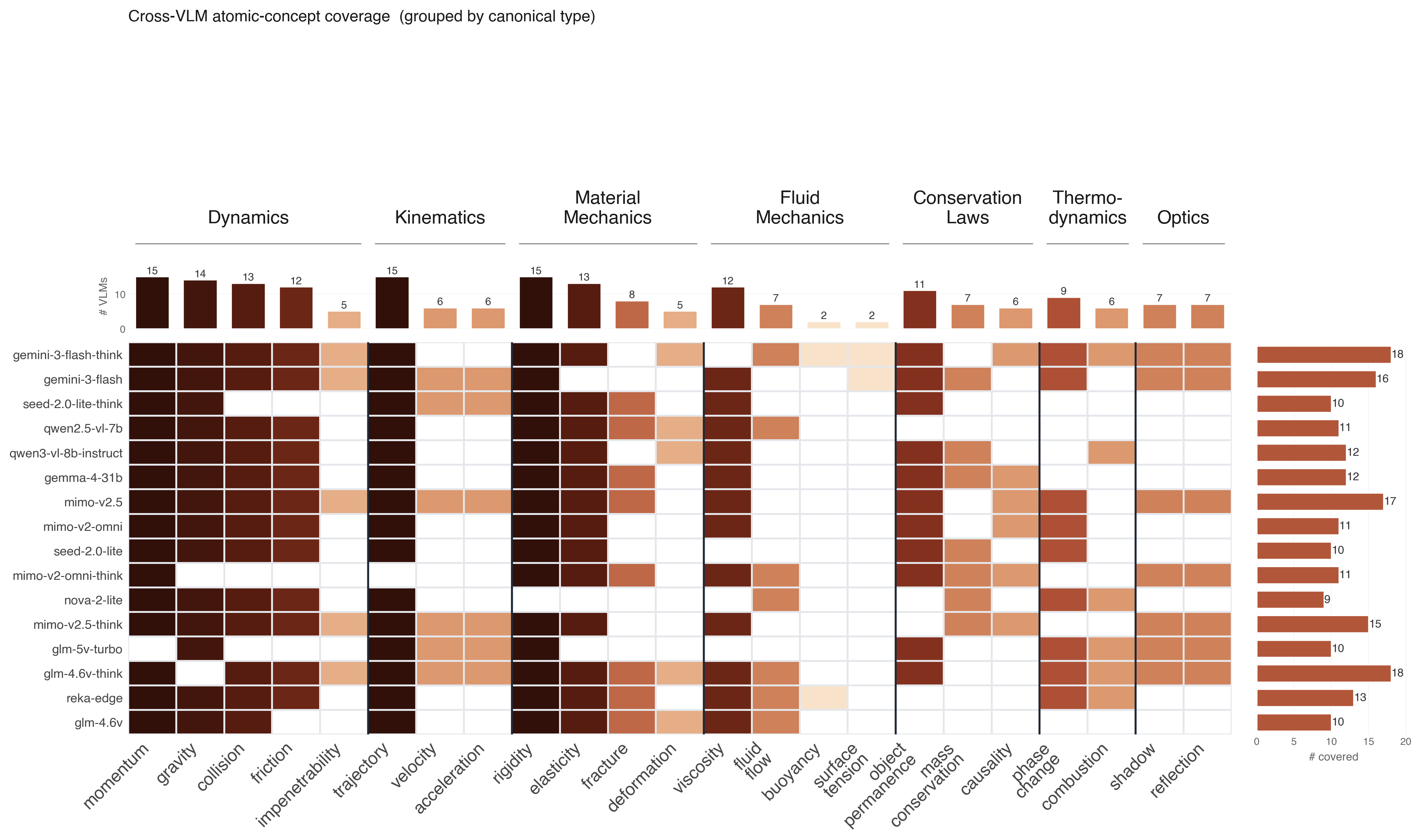}
  \vspace{-2 em}
  \caption{\textbf{Cross-VLM atomic-concept coverage.} Binary coverage matrix over 16 VLMs (rows, ordered by refinement performance) and atomic physical concepts (columns), grouped into canonical types by vertical separators. A cell is filled when some dimension in a VLM's refined taxonomy references that concept. Top marginal: number of VLMs covering each concept; right marginal: number of concepts each VLM covers. A shared core in dynamics, kinematics, and material mechanics is near-universal, while fluid mechanics forms a sparsely covered long tail.}
  \vspace{-1.5 em}
  \label{fig:atomic}
  \vspace{1 em}
\end{figure*}

\subsection{Convergence and Generalization}
\label{sec:convergence}

Figure~\ref{fig:ranking} traces the Spearman correlation $\rho$ on $\mathcal{D}_{\text{annot}}$ across refinement rounds, taken at each round's cumulative-best taxonomy. The mean rises from 0.20 at seed to 0.27, 0.29, and 0.31 over the three rounds, with per-round gains of $+0.07$, $+0.02$, and $+0.02$. The marginal return shrinks quickly: most of the gain arrives in the first round, so a small fixed budget suffices, unlike scalar-reward optimizers that need hundreds of iterations.

The judges differ sharply in how much their seed captures, since it reflects only the errors a VLM reports on its own. A judge that mentions few therefore ends up with a thin taxonomy. Refinement addresses this through the LLM editor, which reads the human ratings and residuals and revises the taxonomy to name the dimensions the VLM never raised on its own, prompting it to score what it had left unspoken. Reka-Edge surfaces the least at seed, yet a single round recovers it to a competitive level, the largest gain in the study.

Finally, the gains transfer from the annotation set to held-out videos. 
Refinement selects taxonomies by the objective on $\mathcal{D}_{\text{annot}}$, yet the resulting improvements over \textsc{PhyGen} hold on the disjoint test set (Table~\ref{tab:vlm_results}) for every model. Because the taxonomy is a short, named set of dimensions rather than a high-capacity predictor, it has little room to overfit the 200 annotation videos, and the calibrated agreement it yields generalizes rather than memorizing the alignment target.

\section{Profiling VLM Judges}
\label{sec:profiling_vlms}

Having established that refinement improves agreement for every judge, we now use the refined per-VLM taxonomies as a lens on the judges themselves, asking which physical concepts they share and how they differ in using the error scale.

\subsection{Cross-VLM Concept Convergence}
\label{sec:concept_convergence}

The refined taxonomies are discovered independently per VLM, raising the question of whether judges converge on a shared set of physical concepts or fragment into model-specific vocabularies. To examine this, we pool the dimensions of all 16 refined taxonomies (64 dimensions in total) and cluster them with Gemini-2.5-Pro into 24 atomic physical concepts grouped under 8 canonical types from classical mechanics. Each dimension is assigned to every atomic concept its definition references, and a VLM is said to cover a concept if any of its refined dimensions maps to it. Figure~\ref{fig:atomic} shows the resulting binary coverage matrix, with VLMs ordered by refinement performance and concepts grouped by type.

Coverage is broadly shared rather than idiosyncratic: 21 of the 24 concepts (87.5\%) appear in at least three VLMs, and no concept is unique to a single judge. A small core is near-universal, with momentum, trajectory, and rigidity each covered by 15 of 16 VLMs, followed by gravity (14), collision, and elasticity (13). These are the concepts every judge independently arrives at, and they cluster in dynamics, kinematics, and material mechanics, where motion and contact are directly visible.

The long tail is where judges diverge. Buoyancy and surface tension are covered by only 2 of 16 VLMs each, and deformation and impenetrability by 5 each. Fluid mechanics is the weakest type overall, with three of its four concepts in this tail, indicating that physical phenomena requiring fine-grained perception of material and fluid behavior surface in few judges' taxonomies. Breadth itself varies widely, from 18 concepts down to a median of 12, and notably does not track refinement performance: GLM-4.6V-Thinking and Gemini-3-Flash-Thinking reference equally many concepts (18 each) yet sit at opposite ends of Table~\ref{tab:vlm_results}. A judge can name many physical phenomena without scoring any reliably against human ratings, the gap refinement closes.

\begin{figure}[t]
  \includegraphics[width=\linewidth]{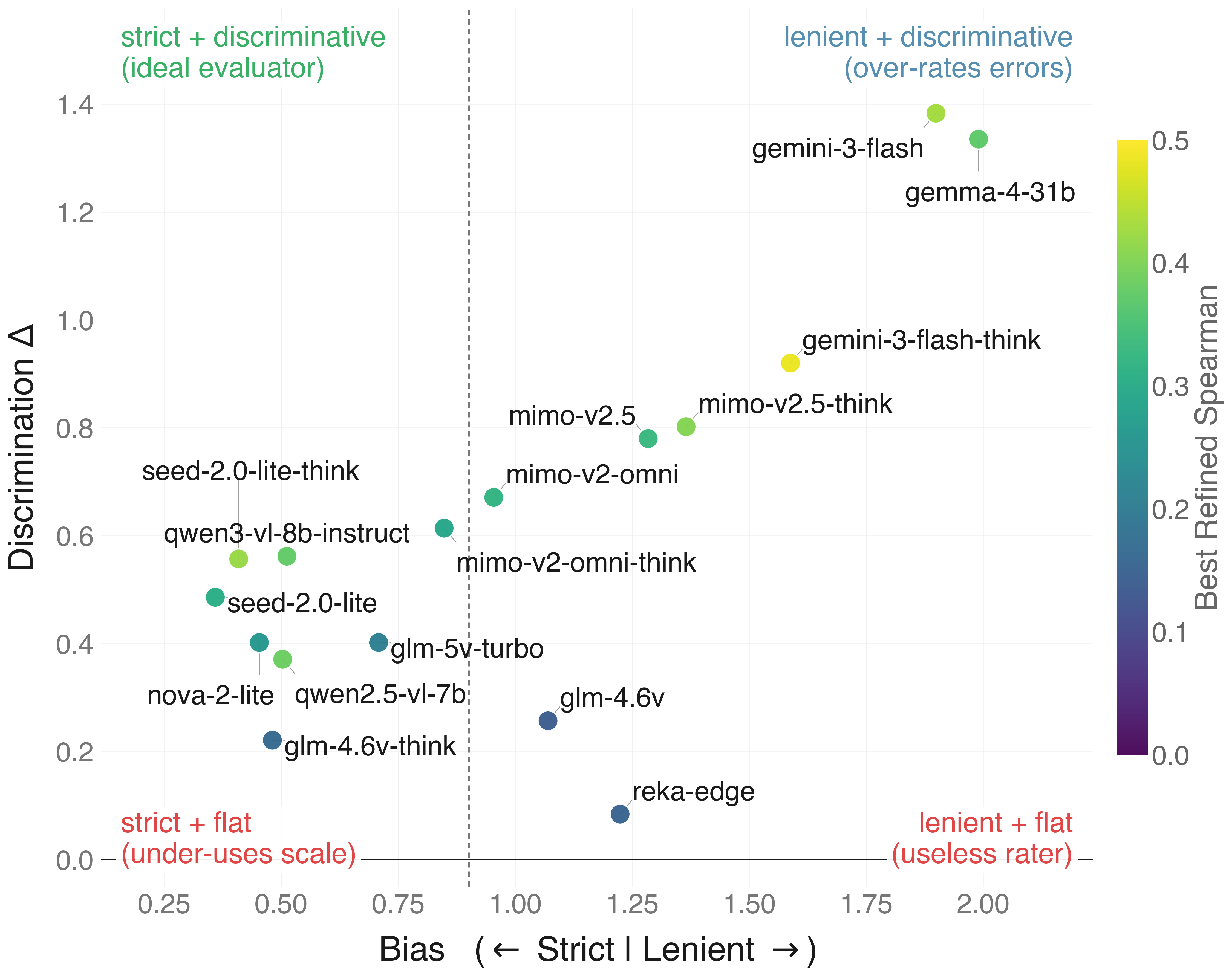}
  \vspace{-2 em}
  \caption{\textbf{Bias-by-discrimination typology of the 16 VLMs.} Each VLM is placed by its bias (x; left strict, right lenient) and discrimination $\Delta$ (y), colored by refined Spearman $\rho$. The dashed line marks the median bias, splitting strict from lenient.}
  \vspace{-1 em}
  \label{fig:bias}
\end{figure}

\subsection{Bias and Discrimination Across Judges}
\label{sec:bias_discrim}

Concept coverage describes which physical phenomena a judge attends to, not how it uses its error scale once it does. We characterize each judge along two scoring traits measured on the test set. \emph{\textbf{Bias}} is the mean error severity it assigns across all videos and dimensions, with low values marking a strict judge and high values a lenient one. \emph{\textbf{Discrimination}} is $\Delta = \bar{s}_{\text{pc}=1} - \bar{s}_{\text{pc}=5}$ \citep{embretson2025item}, the gap between the error severity it assigns to the worst and best videos, with positive $\Delta$ meaning it rates implausible videos higher in error than faithful ones. Splitting bias at its median over the 16 judges gives a four-quadrant typology, shown in Figure~\ref{fig:bias}.

All 16 judges discriminate in the correct direction ($\Delta > 0$): none flatten or invert the human ordering, leaving the bottom two quadrants empty. Judges split cleanly along the bias axis instead, 8 strict and 8 lenient, differing in scale placement, not discrimination.

Bias and discrimination are coupled: more lenient judges also draw wider gaps between good and bad videos, since a judge that freely assigns high error counts has room to separate worst from best. This is why the two lower quadrants stay empty, as a strict judge compresses its scores and cannot also score flat. But bias does not track agreement with human ratings: the strict and lenient halves reach almost the same mean refined correlation, and each spans nearly the full range, from Seed-2.0-Lite-Thinking (0.422) to GLM-4.6V-Thinking (0.163) among strict judges alone. How a judge places its scale is a stable trait of its style, separate from how well it aligns with human judgment.

\section{Conclusion}

We treated a judge's evaluation schema as a property of the evaluator and introduced \textsc{JudgeFit}, which discovers a per-VLM taxonomy by grounding it in each judge's own error vocabulary and refining it under a calibrated diagnosis of what the judge can reliably score. Across 16 VLMs from eight families, it improves agreement with human ratings over a fixed schema for every judge, by 32\% on average, within a small budget. Beyond aggregate accuracy, the per-VLM profiles reveal structure that overall rankings cannot capture, exposing where judges converge, where they diverge, and how their scoring styles vary independently of how well they align with human judgment.

\section*{Limitations}
Although \textsc{JudgeFit} yields consistent gains across all evaluated VLMs, two limitations remain. First, it requires human-annotated data carrying both ratings and per-video problem annotations, yet it needs far fewer such annotations than existing approaches and thus keeps annotation cost low. Second, the seed and refine stages call an LLM, whose reproducibility is not guaranteed. This is inherent to LLM-based pipelines, but its randomness does not affect deployment: once a better taxonomy is found it can be applied directly, and the resulting scoring no longer involves the LLM. We leave validation across a broader range of settings to future work.

% \section*{Ethics Statement}

\bibliography{custom}

\appendix

\section{Implementation Details}
\label{app:impl}

\subsection{Evaluated VLMs.}
\label{app:vlms}
We assess 16 VLMs that natively accept video input, spanning eight families. From Google we include Gemini-3-Flash and its thinking variant Gemini-3-Flash-Thinking \citep{google2025gemini3flash}, and the open-weight Gemma-4-31B \citep{gemma4_2026}. From ByteDance we include Seed-2.0-Lite and Seed-2.0-Lite-Thinking \citep{seed2026seed2}. From Xiaomi we include MiMo-v2.5, MiMo-v2.5-Thinking, MiMo-v2-Omni, and MiMo-v2-Omni-Thinking \citep{xiao2026mimov2flash}. From Alibaba we include Qwen2.5-VL-7B \citep{bai2025qwen25vl} and Qwen3-VL-8B-Instruct \citep{qwen2025qwen3vl}. From Zhipu we include GLM-4.6V, GLM-4.6V-Thinking \citep{glmv2025}, and GLM-5V-Turbo \citep{zai2026glm5v}. We further include Amazon Nova-2-Lite \citep{amazon2025nova2} and Reka-Edge \citep{ormazabal2024reka}. The two Qwen judges are run locally; all others are accessed through OpenRouter. The LLM editor and clustering model are both Gemini-2.5-Pro \citep{geminiteam2025gemini25}, chosen for its long context window, which the clustering step requires to pool free-form error descriptions over the full annotation set.

Several VLMs were considered but not benchmarked. Some do not accept native video input and were excluded outright. Others accept video yet do not produce usable judgments under our protocol: LLaVA-OneVision-7B \citep{li2024llava} returns an all-zero error profile for every clip, InternVL3-8B \citep{zhu2025internvl3} reports no errors on any video, and Qwen3-VL-8B-Thinking emits unbounded reasoning text that fails to complete the annotation set. We exclude all of these rather than report unreliable scores.

\begin{figure*}[t]
\centering
\begin{tcolorbox}[colback=gray!5,colframe=gray!50,boxsep=4pt,arc=2pt,
                  title={LLM editor prompt (refine stage), abbreviated}]
\small\ttfamily
You are refining a physics error taxonomy by proposing local edit operations.
The taxonomy is scored by how well its per-dimension VLM scores predict human
physical-commonsense ratings (pc, 1--5), via a 5-fold cross-validated ridge
regression. Aim for high positive spearman(pc) together with a positive
specificity gain over a one-feature halo baseline that uses only the mean
dimension score.

\vspace{0.4em}
\textbf{\#\# Current Taxonomy}\quad\textit{\{N dimensions, each with a one-line definition\}}\\
\textbf{\#\# Diagnostic Report}\quad\textit{\{global metrics; per-dimension $(\beta_i,\,\beta^{\text{cv}}_i,\,\pi_i,\,\rho_i,\,\text{verdict})$; pairwise correlation matrix with redundant pairs $|r_{ij}|>\gamma$; top mispredicted videos with residual, prompt, VLM open-annotation errors, violated rules, and per-dim scores\}}\\
% \textbf{\#\# Operation History}\quad\textit{\{prior rounds' candidates and metrics, with the current best marked\}}
\vspace{0.4em}
\textbf{\#\# Verdict definitions}\\
$\bullet$ \textsc{noise}: $|\rho_i| < \tau$ — no rank signal against pc.\\
$\bullet$ \textsc{unstable}: $\pi_i < \kappa$ — coefficient sign flips across folds.\\
$\bullet$ \textsc{predictive}: $|\beta_i|$ large, low cross-fold variance, consistent sign.\\
$\bullet$ \textsc{weak}: some signal but $\beta_i$ too small or unstable to be predictive.

\vspace{0.4em}
\textbf{\#\# Available Operations}\\
$\bullet$ \textsc{merge}(dim\_a, dim\_b) $\to$ (new\_name, new\_definition)\\
$\bullet$ \textsc{split}(dim) $\to$ (sub\_1, def\_1), (sub\_2, def\_2)\\
$\bullet$ \textsc{redefine}(dim) $\to$ new\_definition\\
$\bullet$ \textsc{add}(name, definition)\\
$\bullet$ \textsc{drop}(dim)\\
$\bullet$ \textsc{no\_op}

\vspace{0.4em}
\textbf{\#\# Rules} (excerpt)\\
% $\bullet$ Every operation must cite specific diagnostic evidence (verdict, $\beta_i$, residual videos, or violated-rule patterns).\\
$\bullet$ \textsc{drop} only when verdict is \textsc{noise}; for \textsc{unstable} or \textsc{weak} dimensions prefer \textsc{redefine}.\\
$\bullet$ \textsc{merge} when a pairwise correlation exceeds $\gamma$; alternatively \textsc{redefine} one dimension to disambiguate.\\
$\bullet$ \textsc{add} or \textsc{split} when residual videos concentrate on a physical pattern no dimension covers.\\
$\bullet$ Keep 3--8 dimensions, each grounded in physics principles and describing an observable class of violations rather than abstract or subjective qualities.\\
$\bullet$ Maximize the selection objective $J = \rho_{\text{pc}} + \Delta_{\text{spec}}$ (Eq.~\ref{eq:objective}); a candidate that raises $\rho_{\text{pc}}$ by collapsing the taxonomy toward the halo baseline does not win.\\
$\bullet$ Propose exactly $m=2$ independent candidates.

\vspace{0.4em}
\textbf{Return JSON}:
\begin{verbatim}
{"candidates": [
  {"operation": "MERGE",
   "args": {"dim_a": "...", "dim_b": "...",
            "new_name": "...", "new_definition": "..."},
   "reasoning": "cite verdict / beta / residual evidence"},
  {"operation": "ADD",
   "args": {"name": "...", "definition": "..."},
   "reasoning": "cite residual videos / unfollowed rules"}
]}
\end{verbatim}
\end{tcolorbox}
\vspace{-1 em}
\caption{LLM editor prompt used in the refine stage. Fields in braces are filled per VLM and per round.}
\label{fig:editor_prompt}
\end{figure*}

\subsection{Hyper-parameters and Cost}
\label{app:hparams}

\noindent\textbf{Hyper-parameters selection}
Each VLM judge ingests the full video clip and scores every dimension on a $[0,5]$ severity scale. To reduce sampling noise, each (video, dimension) pair is scored $3$ times and averaged, with decoding temperature $0$, and all judges receive an identical prompt template; only the taxonomy supplied to the judge differs across conditions. The diagnostic map $f_{\mathcal{T}}$ is a ridge regression from standardized per-dimension scores to human ratings with regularization strength $\lambda = 1.0$, and the coefficient $\beta_i$, sign consistency $\pi_i$, and rank correlations $\rho_i, r_{ij}$ are computed under $5$-fold cross-validation, so no video used to fit the calibrator contributes to its own diagnosis. A dimension is flagged for \textsc{drop} when $|\rho_i| < \tau$ ($\tau = 0.10$), for \textsc{redefine} when $\pi_i < \kappa$ ($\kappa = 0.6$), and a pair for \textsc{merge} when $r_{ij} > \gamma$ ($\gamma = 0.6$), with these thresholds fixed across all 16 judges. In the seed stage, each judge produces free-form error descriptions over the full $200$-video $\mathcal{D}_{\text{annot}}$, which the clustering LLM groups into an initial taxonomy of roughly four dimensions (range $3$--$7$); in refinement, the editor proposes $m = 2$ candidates per round, each accepted only if it improves the selection objective $J$, up to a budget of $T_{\max} = 3$ rounds. The full editor prompt is shown in \Cref{fig:editor_prompt}.

\noindent\textbf{Compute and cost.}
The two Qwen models are run locally on NVIDIA A100 GPUs (approximately 100 GPU-hours in total); all other judges, together with the LLM editor and clustering model, are accessed through the OpenRouter API. The total API cost, including reruns and excluded models, is approximately \$1500.

\end{document}